\documentclass{article} 
\usepackage{iclr2016_conference,times}
\usepackage{hyperref}
\usepackage{url}
\usepackage{graphicx}
\usepackage{multirow}
\usepackage{algorithm}
\usepackage{algpseudocode}

\title{Similarity-based text recognition by \\Deeply Supervised Siamese Network}

\author{Ehsan~Hosseini-Asl\thanks{ The work was performed during an internship at Captricity, Inc.} \\
	Electrical and Computer Engineering\\
	University of Louisville\\
	Louisville, KY 40292, USA \\
	\texttt{ehsan.hosseiniasl@louisville.edu} \\
	\And
	Angshuman Guha \\
	Captricity, Inc.\\
	Oakland, CA 94612 USA  \\
	\texttt{angshumang@captricity.com} \\
}

\begin{document}

	\maketitle
	
	\begin{abstract}
		In this paper, we propose a new text recognition model based on measuring the visual similarity of text and predicting the content of unlabeled texts. First a Siamese convolutional network is trained with deep supervision on a labeled training dataset. This network projects texts into a similarity manifold. The Deeply Supervised Siamese network learns visual similarity of texts. Then a K-nearest neighbor classifier is used to predict unlabeled text based on similarity distance to labeled texts. The performance of the model is evaluated on three datasets of machine-print and hand-written text combined.  We demonstrate that the model reduces the cost of human estimation by $50\%-85\%$.  The error of the system is less than $0.5\%$. The proposed model outperform conventional Siamese network by finding visually-similar barely-readable and readable text, e.g. machine-printed, handwritten, due to deep supervision. The results also demonstrate that the predicted labels are sometimes better than human labels e.g. spelling correction.
	\end{abstract}
	
	\section{Introduction}
	
	Optical Character Recognition (OCR) is traditionally used to convert images of machine printed text into textual content.  Intelligent Character Recognition (ICR) is used to do the same for images of handwritten text.  State-of-the-art OCR engines can work well, but only for clean data and where the OCR engine can be adjusted to deal with a single font or a small set of fonts.  State-of-the-art ICR engines are significantly worse.  
	
	For a real-life application of high-accuracy character recognition involving both machine print and handwriting, one has to develop one{'}s own OCR/ICR engine.  This \textit{typically} requires plenty of character-segmented data, as well as labeling at the character level.  This is a very expensive proposition in most real-world situations, if not an impossible one.  To avoid the character segmentation cost, \cite{keeler1991self} proposed learning character segmentation and recognition simultaneously from un-segmented data.  This does not work well in practice beyond numeric characters and for large vocabularies.  There has been some work at limited-vocabulary whole word recognition, see, for example, ~\cite{lavrenko2004holistic}.  To avoid character segmentation, \cite{lecun1998gradient} proposed a graph transformer network with Viterbi search.  These kinds of models cannot compete in performance (training time) with modern deep neural nets that afford efficient implementations, for example, using GPU.  \cite{bunke2004offline} handled the segmentation problem using HMM-based recognition using the Viterbi algorithm.  The present authors have experienced (convolutional) neural nets consistently out-performing HMMs in at least two domains, both with large quantities of industrial data:  online handwriting recognition and speech recognition.  Recently there also has been work involving innovative models where $n^{th}$ characters are predicted for an input word of a fixed-size input image.  For instance, one model might predict the first character, another model might predict the second character, and so on.  See for example,~\cite{jaderberg2014synthetic}.
	
	\begin{figure}[htb!]
		\begin{center}
			\includegraphics[width=1\linewidth]{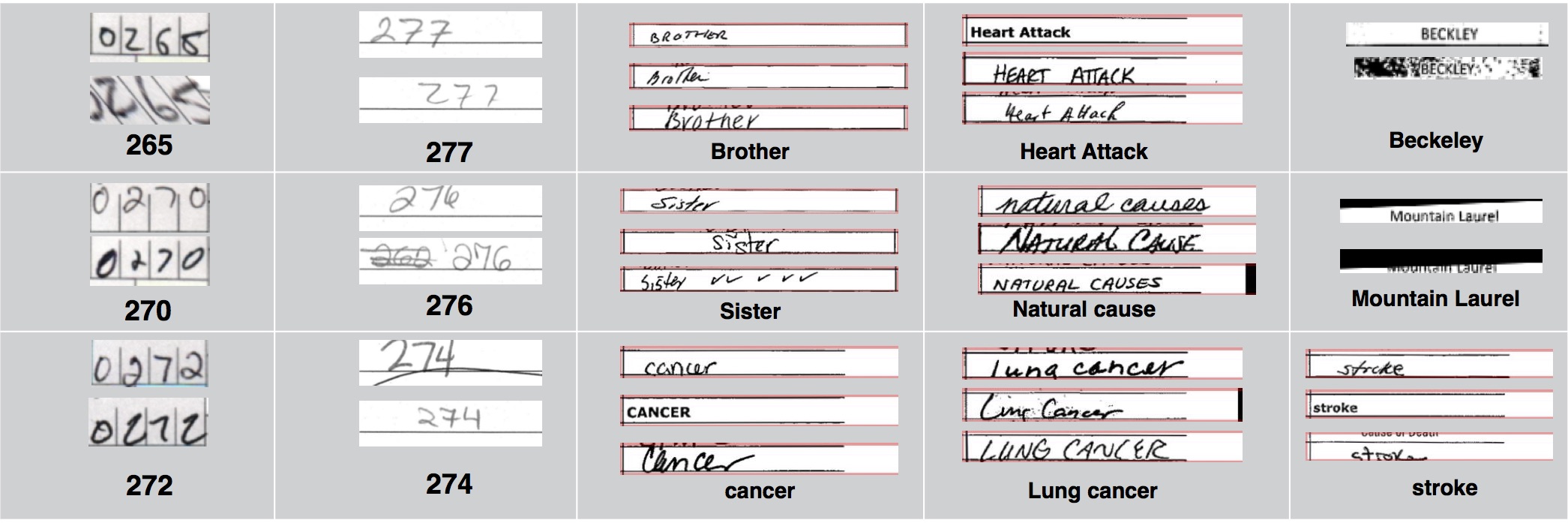}
		\end{center}
		\caption{Examples of similar text.}
		\label{fig:similar-text}
		\vspace{-3mm}
	\end{figure}

	Our goal is for a practical system that organically incorporates human labeling with machine learning to achieve very low error rates ($\le0.5\%$) while minimizing the amount of necessary human labeling.
	
	Siamese Network (SN) is a type of end-to-end metric learning approach, consisting of a neural network that learns a discriminative function. SNs are trained by learning a similarity metric between pairs of data.  SN models are applied to signature verification~\cite{bromley1993signature}, digit recognition~\cite{hadsell2006dimensionality}, face recognition~\cite{chopra2005learning},  Speech feature extraction~\cite{chen2011extracting}, and Speech keyword detection~\cite{grangier2007learning}. In this paper, we propose and discuss a novel method of text recognition that does not require character-segmented data. by predicting the content of a text using similar labeled texts.  We use a Siamese Convolutional Network to learn the similarity between text images in a low-dimensional Euclidean space. To account for similar machine-printed and handwritten text, the SN is regularized by supervision in hidden layers~\cite{weston2012deep,lee2014deeply}. Then a k-nearest neighbor algorithm is employed to predict the label based on most similar labeled texts. To account for unseen labels in test data, an interactive k-nearest neighbor algorithm with human annotation is employed for label prediction, and reducing the error.  See Fig.~\ref{fig:similar-text} for examples of “similar” images i.e. with the same text content.

	\section{Proposed Model}
	In this section, we propose a model of text recognition based on learning similarity of images. In section~\ref{sec:model}, a Siamese network is proposed for learning similarity of text, and section~\ref{sec:framework} describes a text recognition framework.

	\subsection{Learning Text Similarity}
	\label{sec:model}
	To train a model to be able to learn the similarity between texts, a Siamese network is used as in~\cite{chopra2005learning, hadsell2006dimensionality}.  The Siamese network is trained to project the images into a feature space, where similar images are projected with short mutual Euclidean distance, and dissimilar images are projected with large mutual Euclidean distances. Training of the Siamese network is based on minimizing the contrastive loss of a pair of images,
	\begin{equation}
		\mathcal{L}(W)=(1-Y)*\frac{1}{2}D_{w}^{2}+\frac{1}{2}*Y*{max(0,m-D_{w})^{2}}
		\label{eq:loss}
	\end{equation}
	
	\begin{figure}[b]
		\begin{center}
			\includegraphics[width=1\linewidth]{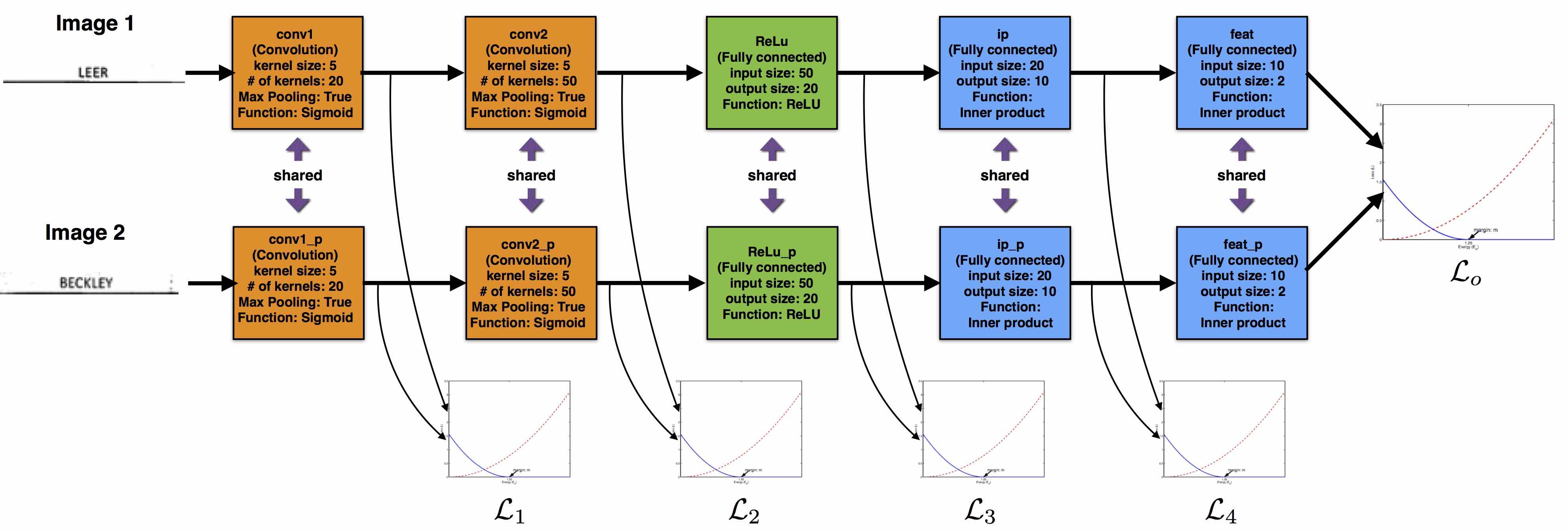}
		\end{center}
		\caption{Deeply Supervised Siamese Network (DSSN) for learning text similarity}
		\label{fig:model}
	\end{figure}

	where $W=\{\{w^{0},...,w^{n}\},w^{o}\}$ are the weights of the hidden layers and output layer of the Siamese network, $Y$ is the label of paired images, i.e. 0 if similar and 1 if dissimilar, $D_{w}$ is the Euclidean distance between a pair of images, and $m$ is the desired Euclidean distance between a pair of dissimilar images. Siamese networks have shown promising results in learning similarity of the handwritten digits dataset, MNIST.  However, in complicated cases of long text, capturing similarities between two texts is infeasible using a single loss function in the output layer of Siamese network. The performance of contrastive loss $L$ is dependent on feature extraction of the hidden layers, where it should capture the similarities in a hierarchical way, to enable the output layer to extract features which can clearly represent the similarities of long and complex text. In order to boost the performance of the Siamese network for learning similarity of long text, we used the method of deep supervision proposed in~\cite{lee2014deeply}, where several contrastive loss functions are used for hidden and output layers, to improve the discriminativeness of feature learning, as shown in Fig.~\ref{fig:model}. Therefore, the proposed method is called Deeply Supervised Siamese Network (DSSN) and it is trained using the combined contrastive loss,
	
	\begin{equation}
	\mathcal{L}_{DSSN}(W)=\sum_{l=0}^{n}\mathcal{L}_{l}(w^{(l)}) + \mathcal{L}_{o}(w^{o})
	\label{eq:d-loss}
	\end{equation} 
    where $l$ indicates the index for hidden layer, $o$ is the output layer. Eq.~\ref{eq:d-loss} indicates that the loss $\mathcal{L}_{l}$ of each hidden layer is only the function of weights of that layer, i.e. $w^{(l)}$. The DSSN network generates a Similarity Manifold, where similar texts are projected with short mutual Euclidean distances. The next section describes the text recognition model based on the Similarity Manifold. The ADADELTA method of gradient descent (\cite{zeiler2012adadelta}) is used to update the parameters of DSSN.

	\subsection{Text Recognition by Text Similarity}
	\label{sec:framework}
\begin{figure}[t]
		\begin{minipage}[b]{1\linewidth}
			\centering
			\includegraphics[width=12cm]{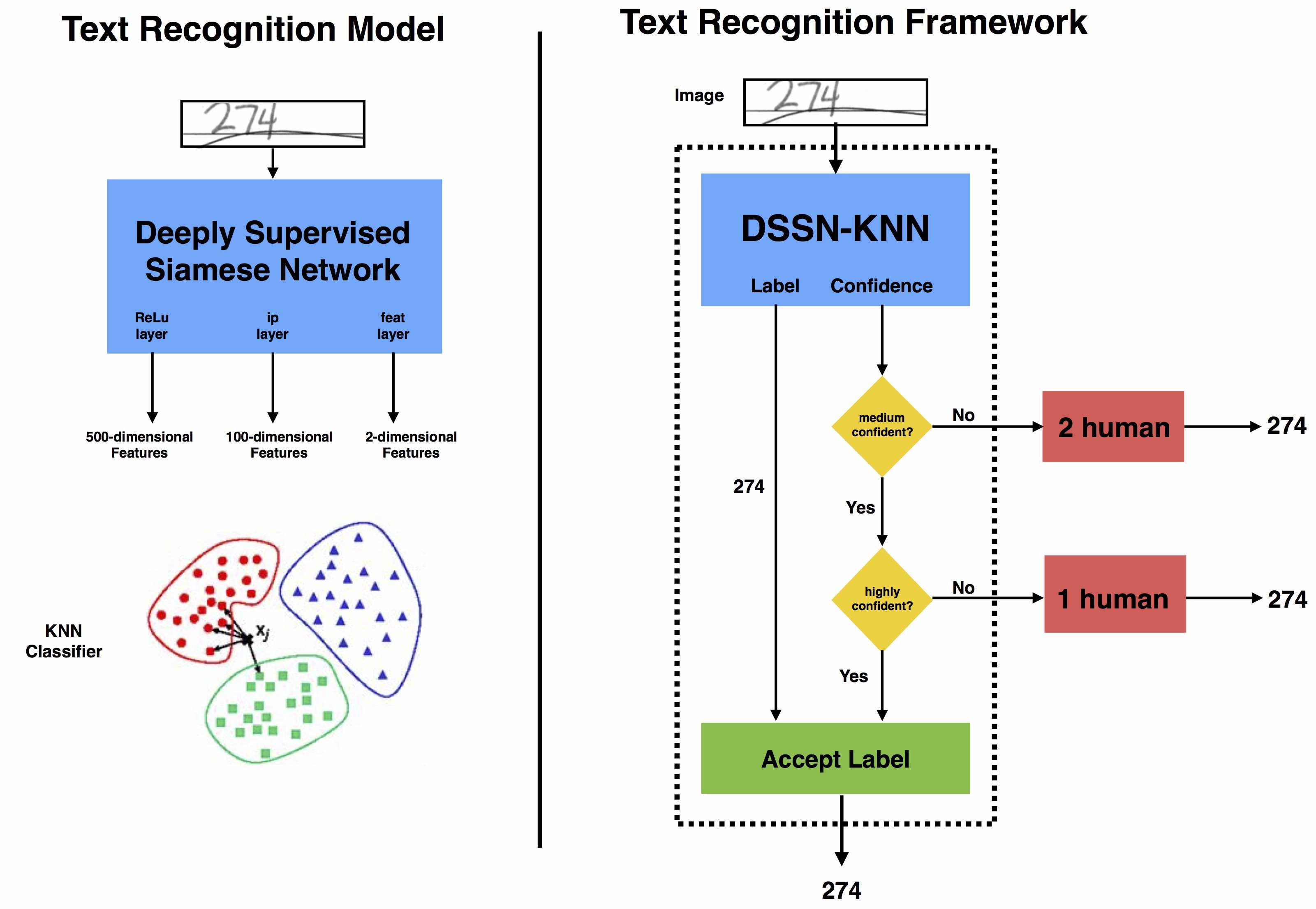}
			
			{{\footnotesize (a)}}
		\end{minipage}
		
		\begin{minipage}[b]{1\linewidth}
			\centering
			\includegraphics[width=10cm]{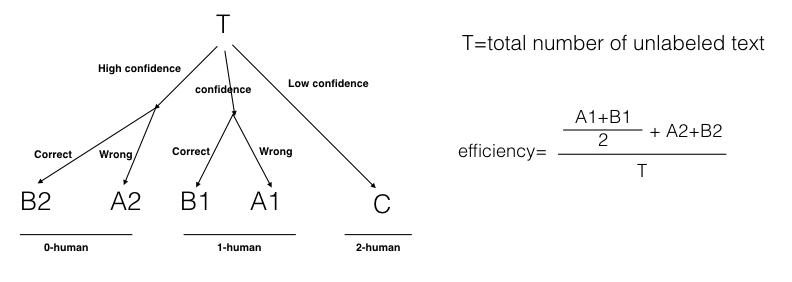}
			
			{{\footnotesize (b)}}
		\end{minipage}
		\caption{(a) Proposed Framework for text recognition based on text similarity. (b) Efficiency metric to measure the reduction in human estimation of label.}
		\label{fig:framework}
		\vspace{-3mm}
	\end{figure}

	This section describes a text recognition framework to predict the label of text using the DSSN model developed in the previous section. The text recognition model is based on feature extraction of text using proposed DSSN network. We use a K-nearest neighbor algorithm to predict the label of text images in test data, based on similarity distance to the labeled text in train data. The predicted label is compared with human estimation as shown in Fig.~\ref{fig:framework}(a). 
	
    Our human-based model for text label prediction is based on voting of two humans on a text image. The proposed framework in Fig.~\ref{fig:framework} (a) is motivated by our goal of reducing the cost of human estimations while maintaining a low error rate. As shown in Fig.~\ref{fig:framework}(a), the predicted label of DSSN-KNN is accompanied by a \textit{confidence} value.  We choose two parameters, $\theta_{1}$ and $\theta_{2}$, such that the confidence value can be classified as \textit{highly confident}, \textit{confident} and \textit{not confident}.  If the model's prediction confidence is high, we omit the required two human estimations. However, if the prediction is only confident, we validate the predicted label of DSSN-KNN with one human estimation.  The parameters $\theta_{1}$ and $\theta_{2}$ are chosen by tuning the model's performance on the training set (or one can use a validation set).
	
	To measure the performance of DSSN-KNN in reducing the human estimation, we define an efficiency metric as shown in Fig.~\ref{fig:framework}(b),
	
	\begin{equation}
		\mathtt{efficiency}=\frac{\frac{A_{1}+B_{1}}{2}+A_{2}+B_{2}}{T}
	\end{equation}
	where $T$ is the total number of text samples, $A_{1}$ and $B_{1}$ are the number of confidently wrong and confidently correct predictions, and $A_{2}$ and $B_{2}$ are the number of high-confident wrong and high-confident correct predictions, respectively.  

    Note that the efficiency metric definition implicitly assumes a low rate of disagreement between two humans labeling the same image or between a human and the DSSN-KNN model.  If this rate is $1\%$ (which is what we see in practice, see AC column in Table~\ref{tab:thresholds}), the metric will overcount the reduction in the required number of human estimates by $\sim1\%$.  In the case of disagreement, extra human estimates will be needed to resolve conflicts.

    The DSSN-KNN model can be used in one of two modes: ROBOTIC and ASSISTIVE.

    ROBOTIC mode is suggested by Fig.~\ref{fig:framework} -- (i) for high confidence predictions, we skip human labeling, (ii) for medium confidence predictions, we ask for human confirmation and (iii) for low confidence predictions, we discard the prediction and ask for at least two human estimates.  

    ASSISTIVE mode is to ignore $\theta_{2}$ (high confidence threshold) -- (i) for high and medium confidence predictions, we ask for human confirmation and (ii) for low confidence predictions, we discard the prediction and ask for at least two human estimates.

    By definition, ASSISTIVE mode results in zero error from the DSSN-KNN model.   But efficiency is lower because $\{A,B\}_{2}$ are folded into $\{A,B\}_{1}$ in the numerator of Fig.~\ref{fig:framework}(b). On the other hand, ROBOTIC mode has higher efficiency at the cost of some DSSN-KNN errors unchecked by humans.  We want this error to be under 0.5\%.
	
	\begin{algorithm}[htb!]
		\caption{Interactive Text Recognition by DSSN-KNN}
		\textbf{Data:} text\\
		\textbf{Output:} label
		\begin{algorithmic}[1]
			
			\State {Extract the feature of new text image from hidden layer of DSSN}
			\State {Predict the \textit{label} and confidence using k-nearest-neighbor algorithm}
			\If {mode=ROBOTIC}
			\If {$\theta>\theta_{2} $}
			\State{Output=\textit{label}}
			\ElsIf {$\theta_{1}<\theta<\theta_{2} $}
			\State{verify the label with 1 human}
			\If {DSSN-KNN and human disagree}
			\State {get another human label}
			\If {two humans agree}
			\State {Update KNN dictionary}
			\State{Output=\textit{human label}}
			\Else
			\State {Output=\textit{label}}
			\EndIf
			\Else
			\State {Output=\textit{label}}
			\EndIf
			\Else 
			\State{Label the text with 2 human}
			\State {Update KNN dictionary}
			\State{Output=\textit{human label}}
			\EndIf	
			\Else  {\hspace{1mm}// mode=ASSISTIVE}
			\If {$\theta_{1}<\theta$}
			\State{verify the label with 1 human}
			\If {DSSN-KNN and human disagree}
			\State {get another human label}
			\If {two human agree}
			\State {Update KNN dictionary}
			\State{Output=\textit{human label}}
			\Else
			\State {Output=\textit{label}}
			\EndIf
			\Else
			\State {Output=\textit{label}}
			\EndIf
			\Else 
			\State{Label the text with 2 human}
			\State {Update KNN dictionary}
			\State{Output=\textit{human label}}		
			\EndIf
			\EndIf
		\end{algorithmic}
		\label{alg: prediction}
	\end{algorithm}
	
	\vspace{-3mm}
	\section{Experiments}
	\label{sec:experiment}
	\vspace{-2mm}
	In this section, we design several experiments to evaluate the performance of our proposed model of text recognition. The DSSN-KNN model is pretrained on MNIST data and then fine-tuned on the datasets to minimize the loss function of Eq.~\ref{eq:d-loss}. We select minibatch size of $10$ paired texts, containing $5$ similar pairs and $5$ dissimilar pairs, to train the Similarity manifold. We used Caffe, ~\cite{jia2014caffe}, and Theano, ~\cite{Bastien-Theano-2012}, on Amazon EC2 g2.8xlarge instances with GPU GRID K520 for our experiments. First we apply some metrics to evaluate the performance of DSSN in learning the similarity manifold in section~\ref{sec:exp-sim}. Then the performance of DSSN-KNN is evaluated for text recognition of three hand-written text datasets in section~\ref{sec:exp-rec}. 
	
	\subsection{Similarity Manifold Evaluation}
	\label{sec:exp-sim}
	\vspace{-1mm}
	In order to evaluate the performance of proposed DSSN for text recognition, we evaluated the trained similarity manifold for detecting similar and dissimilar texts. For this purpose, we implemented two separate experiments for non-numeric and numeric texts.
	
	The non-numeric dataset contains 8 classes, where two major classes dominate in sample count. We found that most of the human-labeled 'blanks' are actually not blank, and contain some text from the two major classes. This misclassified text in training data hurts the performance of DSSN.
	
	\begin{figure}[t]
		\begin{minipage}[b]{1\linewidth}
			\centering
			\includegraphics[width=12cm]{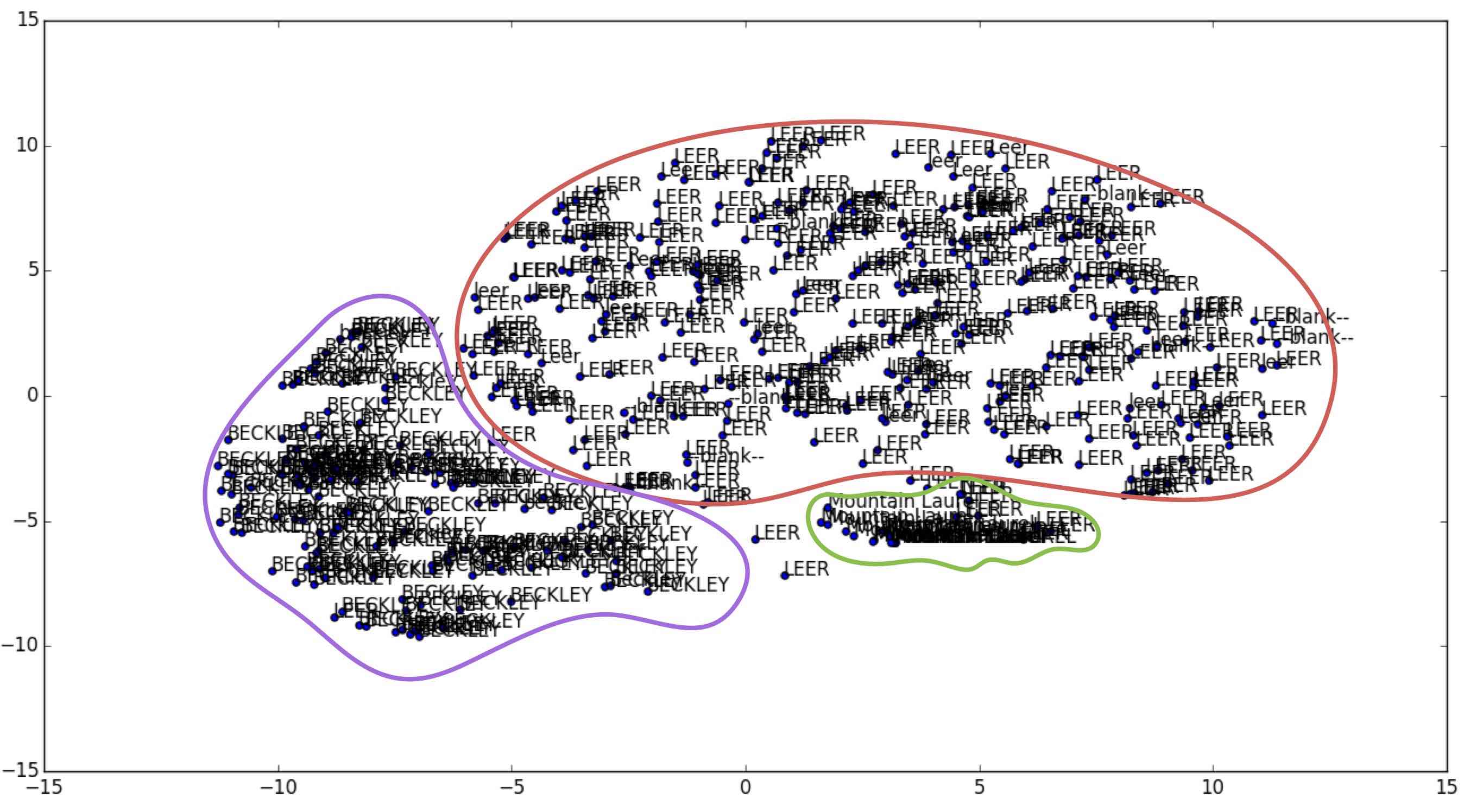}
			
			{{\footnotesize (a)}}
		\end{minipage}
		
		\begin{minipage}[b]{1\linewidth}
			\centering
			\includegraphics[width=9cm]{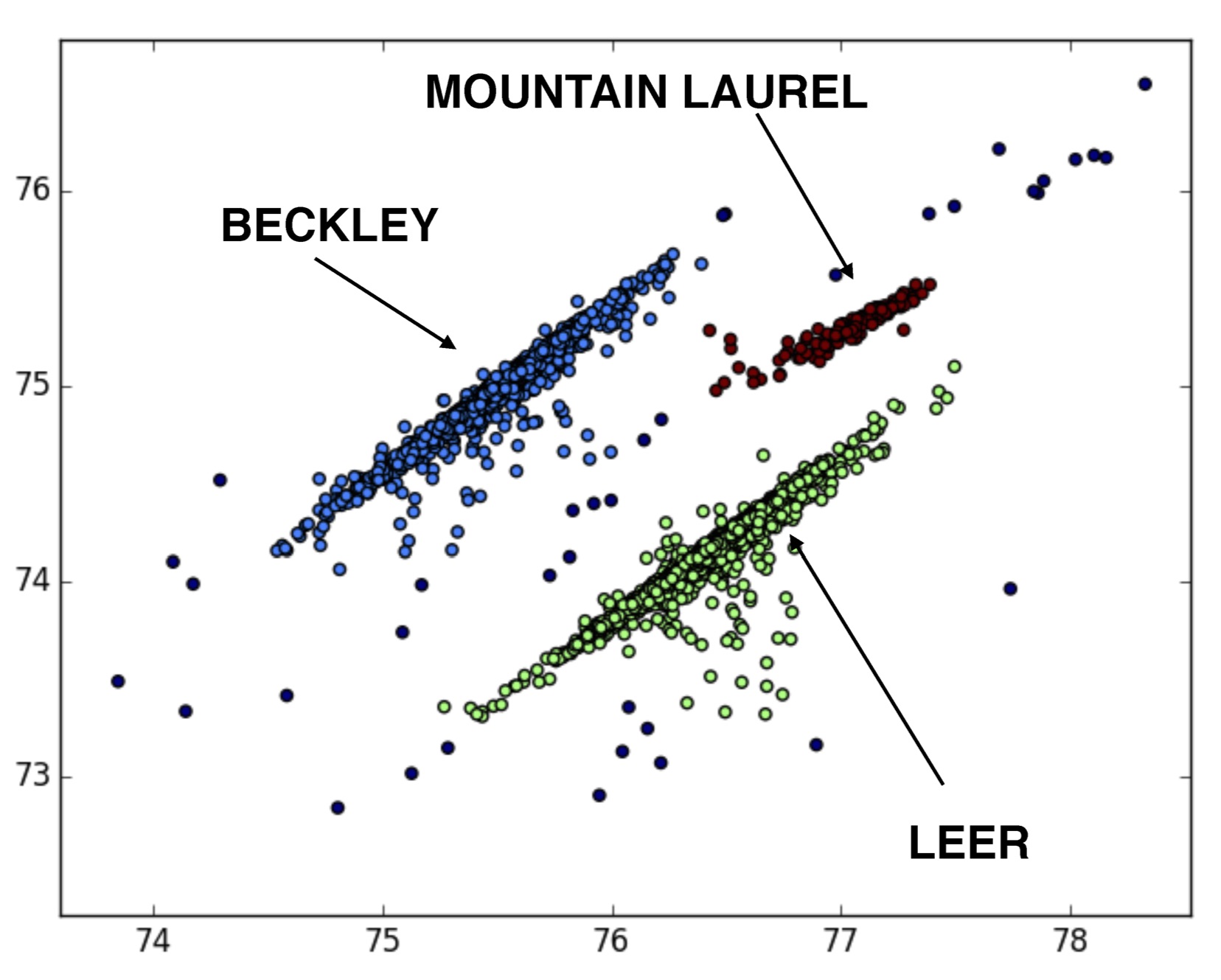}
			
			{{\footnotesize (b)}}
		\end{minipage}
		\caption{Similarity manifold visualization of machine-printed non-numeric texts in (a) hidden and (b) output layer, using t-SNE projection.}
		\label{fig:mp-manifold}
		\vspace{-3mm}
	\end{figure}

\begin{figure}[htb!]
		\includegraphics[width=0.85\linewidth]{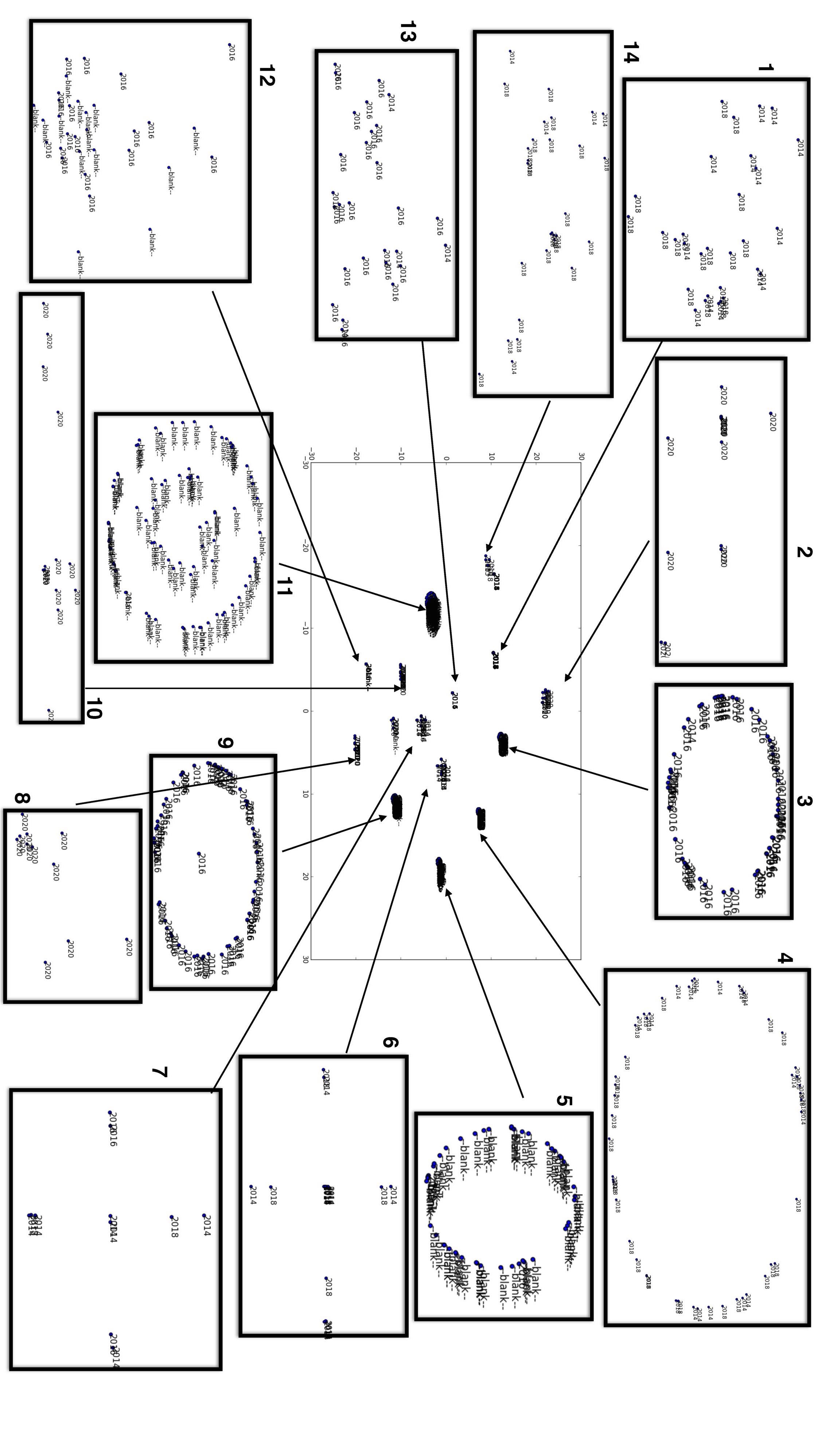}
		\caption{Similarity manifold visualization of machine-printed numeric text using t-SNE projection.}
		\label{fig:mp-numeric-manifold}
	\end{figure}

    To investigate the distribution of text in the similarity manifold, the feature spaces of hidden layers and output layer are visualized in Fig.~\ref{fig:mp-manifold} and Fig.~\ref{fig:mp-numeric-manifold}. Fig.\ref{fig:mp-manifold} shows the visualization of texts based on the 50- and 20-dimensional features extracted in 'conv2' and 'ReLu' layers, respectively. (Visualization of multidimensional data is done using a technique called t-SNE. See \cite{van2008visualizing}.)  It demonstrates that the three major classes are well-separated e.g. 'LEER, ''BECKLEY' and 'Mountain Laurel'. Fig.~\ref{fig:mp-numeric-manifold} depicts the distribution of all texts in 'feat' layer, where each region is expanded for better visualization. Accordingly, some boxes contain texts belonging to only one class, e.g. 2, 3, 5, 8, 9, 10, 11. The '2014' class is mixed with other classes of '2018' and '2016', as shown in boxes 1, 4, 6, 7, 13. The 'blank' shreds in box 12 which are combined with '2016' texts are mis-labeled texts -- reducing the clustering performance of the DSSN model. 
	
In order to evaluate the similarity manifold, several random pairs of images are selected from the test set and feed-forwarded through the DSSN. Then, the Euclidean distance between the paired images is computed based on the output of 'feat' layer. We choose a decision threshold, $\theta$, such that $0.9*FN+0.1*FP$ is minimized over the training set.  Here $FP$ is the false positive rate (similar images predicted as dissimilar) and $FN$ is the false negative rate (dissimilar images predicted as similar). We weigh $FN$ more than $FP$ because the former increases efficiency at the cost of accuracy while the latter does not hurt accuracy.  Table~\ref{tab:sim-error} shows the results for the model initialized by MNIST data, and after fine-tuning on the training dataset.
	
	\begin{table}[htb!]
		\caption{Similarity prediction in Similarity manifold based on Euclidean Distance}
		\label{tab:sim-error}
		\begin{center}
			\begin{tabular}{lccc}
				\multicolumn{1}{c}{\bf DSSN}  &\multicolumn{1}{c}{\bf FN}  &\multicolumn{1}{c}{\bf FN} &\multicolumn{1}{c}{\bf Error}
				\\ \hline \\
				Pretrained by MNIST  & $21.63\%$ &  $7.58\%$ & $14.60\%$ \\
				After Fine-tuning   & $4.61\%$ &  $1.89\%$ & $3.25\%$ \\
			\end{tabular}
		\end{center}
		\vspace{-5mm}
	\end{table}
	
	\begin{table}[htb!]
		\caption{Text Clustering evaluation in Similarity manifold of different layers of DSSN in machine-printed texts}
		\label{tab:sim-clust}
		\begin{center}
			\begin{tabular}{lccc}
				\multicolumn{1}{c}{\bf Dataset Type}  &\multicolumn{3}{c}{\bf Adjusted Rand Index} \\ \multicolumn{1}{c}{} &\multicolumn{1}{c}{\bf feat layer} &\multicolumn{1}{c}{\bf ip layer} &\multicolumn{1}{c}{\bf ReLu layer}
				\\ \hline \\
				non-numeric text  & $0.91$ &  $0.95$ & $0.95$ \\
				numeric text   & $0.96$ &  $0.93$ & $0.96$ \\
			\end{tabular}
		\end{center}
		\vspace{-3mm}
	\end{table}
	
To further evaluate the similarity manifold, a clustering algorithm is applied on texts and the clustered texts are evaluated based on truth labels. For this test, we don't need parallel networks of DSSN.  We use the extracted features from hidden and output layers for clustering of the text. Several clustering algorithms were implemented: K-means, spectral clustering, DBSCAN and agglomerative clustering. To have a better evaluation of features in each layer, we applied clustering algorithms on the features of the 'ReLu', 'ip', and 'feat' layers.  The number of clusters for K-means and spectral clustering were set to 8. For DBSCAN and Agglomerative algorithms, the number of clusters was based on the similarity distance between text samples.  The clustering performance is measured using Adjusted Rand Index (\cite{hub1985comparing}, \cite{rand1971objective}). Table~\ref{tab:sim-clust} shows the best clustering algorithm performance, which was agglomerative clustering on 3 layers of DSSN network.
	

	\subsection{Text Recognition Evaluation}
	\label{sec:exp-rec}
	In section~\ref{sec:exp-sim}, the similarity manifold learned by DSSN was evaluated for clustering and similarity prediction. This section focuses on performance of the proposed DSSN-KNN framework, as shown in Fig.~\ref{fig:framework} for text recognition. The trained DSSN model was tested on three difficult hand-written datasets.  These datasets included hand-written and machine printed text with many variations of translation, scale and image patterns for each class. The number of texts and unique classes in each dataset are listed in Table~\ref{tab:datasets}.
	
	\begin{table}[t]
		\caption{Hand-written text image datasets}
		\label{tab:datasets}
		\begin{center}
			\begin{tabular}{lccc}
				\multicolumn{1}{c}{Dataset\#1 (Short text - Unit)} &\multicolumn{1}{c}{\bf Total data} &\multicolumn{1}{c}{\bf Train Data} &\multicolumn{1}{c}{\bf Test Data}
				\\ \hline \\
				No. of Images  & $90010$ &  $72008$ & $18002$ \\
				No. of Labels   & $1956$ &  $1722$ & $827$ \\
				No. of Unseen Labels   & $11\%$ &  $-$ & $\mathbf{234}$ \\
				No. of blank Images   & $50592$ &  $40517$ & $10075$ \\
			\end{tabular}
		\end{center}
		\begin{center}
			\begin{tabular}{lccc}
				\multicolumn{1}{c}{Dataset\#2 (Short text - Non-Numeric)} &\multicolumn{1}{c}{\bf Total data} &\multicolumn{1}{c}{\bf Train Data} &\multicolumn{1}{c}{\bf Test Data}
				\\ \hline \\
				No. of Images  				& $89580$ &  $71664$ & $17916$ \\
				No. of Labels   			& $1612$ &  $1321$ & $459$ \\
				No. of Unseen Labels   & $18\%$ &  $-$ & $\mathbf{291}$ \\
				No. of blank Images   & $84143$ &  $67309$ & $16834$ \\
			\end{tabular}
		\end{center}
		\begin{center}
			\begin{tabular}{lccc}
				\multicolumn{1}{c}{Dataset\#3 (Short text - Numeric and Non-Numeric)} &\multicolumn{1}{c}{\bf Total data} &\multicolumn{1}{c}{\bf Train Data} &\multicolumn{1}{c}{\bf Test Data}
				\\ \hline \\
				No. of Images  				& $89461$ &  $71568$ & $17893$ \\
				No. of Labels   			& $3124$ &  $2540$ & $792$ \\
				No. of Unseen Labels   & $18.69\%$ &  $-$ & $\mathbf{584}$ \\
				No. of blank Images   & $82864$ &  $66328$ & $16534$ \\
			\end{tabular}
		\end{center}
	\end{table}
	
	The text recognition performance of DSSN-KNN on the three datasets is listed in Table~\ref{tab:final}, where the reduction in human estimation is computed.  The performance of DSSN-KNN is measured by Accuracy (AC), Accuracy of DSSN-KNN High-Confidence predicted labels (HCAC), Accuracy of medium-confident predicted labels validated by a human (HVAC), False Negative labels (FN), and High-Confidence False Negatives (HCFN).  In order to select the confidence and high-confidence thresholds ($\theta_{1}$ and $\theta_{2}$) for each dataset, we did a grid search over the two thresholds to minimize High Confidence False Negative (HCFN). The chosen thresholds for each dataset and the error values are shown in Table~\ref{tab:thresholds}.
	
		\begin{table}[htb!]
			\caption{Text recognition performance on each dataset with respect to $\theta_{1}$ and $\theta_{2}$ to achieve HCFN$\le0.5\%$}
			\label{tab:thresholds}
			\begin{center}
				\scalebox{0.85}{
					\begin{tabular}{llccccccccc}
						\hline\hline
						\multicolumn{1}{c}{Dataset} &\multicolumn{1}{c}{Method} & \multicolumn{1}{c}{Model} &\multicolumn{1}{c}{\bf $\theta_{1}$} &\multicolumn{1}{c}{\bf $\theta_{2}$} &\multicolumn{1}{c}{efficiency} &\multicolumn{1}{c}{AC} &\multicolumn{1}{c}{HCAC} &\multicolumn{1}{c}{ HVAC} &\multicolumn{1}{c}{FN}&\multicolumn{1}{c}{HCFN}
						\\ \hline 
						\multirow{4}{*}{Dataset\#1} &\multirow{2}{*}{ROBOTIC} & SN & $0.94$ &  $0.99$ & $0.48$ & $0.97$ & $0.97$ &$0.27$& $0.01$ & $0.0124$\\
						& & DSSN & $0.94$ &  $0.99$ & $\bf 0.50$ & $\bf 0.99$ & $\bf 0.99$ &$\bf 0.98$& $\bf 0.0049$ & $\bf 0.0033$\\
						& \multirow{2}{*}{ASSISTIVE} & SN	& $0.95$ & $1$& $0.24$& $0.97$ &-&$0.97$&$0.01$& $0$ \\
						& & DSSN	& $0.95$ & $1$& $\bf 0.27$& $\bf 0.99$ &-&$\bf 0.99$&$\bf 0.0047$& $0$ \\
						\hline
						\multirow{4}{*}{Dataset\#2} & \multirow{2}{*}{ROBOTIC} & SN& $0.94$ &  $0.99$ & $0.73$ & $0.79$ & $0.81$ &$0.98$& $0.4461$ & $0.4367$\\ 
						& & DSSN& $0.94$ &  $0.99$ & $\bf 0.87$ & $\bf 0.99$ & $\bf 0.99$ &$\bf 0.98$& $\bf 0.00407$ & $\bf 0.0027$\\ 
						& \multirow{2}{*}{ASSISTIVE} & SN	& $0.95$ & $1$& $0.33$& $0.56$ &-&$0.56$&$0.44$& $0$ \\
						& & DSSN	& $0.95$ & $1$& $\bf 0.45$& $\bf 0.99$ &-&$\bf 0.99$&$\bf 0.0039$& $0$ \\
						\hline
						\multirow{4}{*}{Dataset\#3} & \multirow{2}{*}{ROBOTIC} & SN & $0.94$ &  $0.99$ & $0.82$ & $0.89$ & $0.89$ &$0.85$& $0.11$ & $0.1087$\\
						& & DSSN & $ 0.94$ &  $ 0.99$ & $\bf 0.86$ & $\bf 0.99$ & $\bf 0.99$ &$\bf 0.98$& $\bf 0.0030$ & $\bf 0.0016$\\
						& \multirow{2}{*}{ASSISTIVE} & SN & $0.95$ & $1$& $0.41$& $0.89$ &-&$0.89$&$0.11$& $0$ \\
						& & DSSN & $0.95$ & $1$& $\bf 0.45$& $\bf 0.99$ &-&$\bf 0.99$&$\bf 0.0029$& $0$ \\
						\hline\hline
					\end{tabular}
				}
			\end{center}
		\end{table}

	\begin{table}[htb!]
		\caption{Human-less estimation using proposed DSSN-KNN text recognition model.}
		\label{tab:final}
		\begin{center}
			\begin{tabular}{ccccc}
				\multicolumn{1}{c}{Dataset} &\multicolumn{1}{c}{Type} &\multicolumn{1}{c}{No. of labeled Images} &\multicolumn{2}{c}{\bf Human-less efficiency}\\
				\multicolumn{1}{c}{} &\multicolumn{1}{c}{} &\multicolumn{1}{c}{} &\multicolumn{1}{c}{ROBOTIC}&\multicolumn{1}{c}{ASSISTIVE}
				\\ \hline \\
				Dataset \#1  & machine \& hand &  $18002$ & $8196 / 1659 (50.31\%)$ & $9789 (27.19\%) $\\
				Dataset \#2    & machine \& hand &  $17916$ & $14739 / 1808 (87.31\%)$ & $16475 (45.98\%)$ \\
				Dataset \#3   & machine \& hand &  $17893$ & $14509 / 1706 (85.85\%)$ & $16130 (45.07\%)$ \\
			\end{tabular}
		\end{center}
		\vspace{-5mm}
	\end{table}
	
\begin{figure}[t]
		\includegraphics[width=1\linewidth]{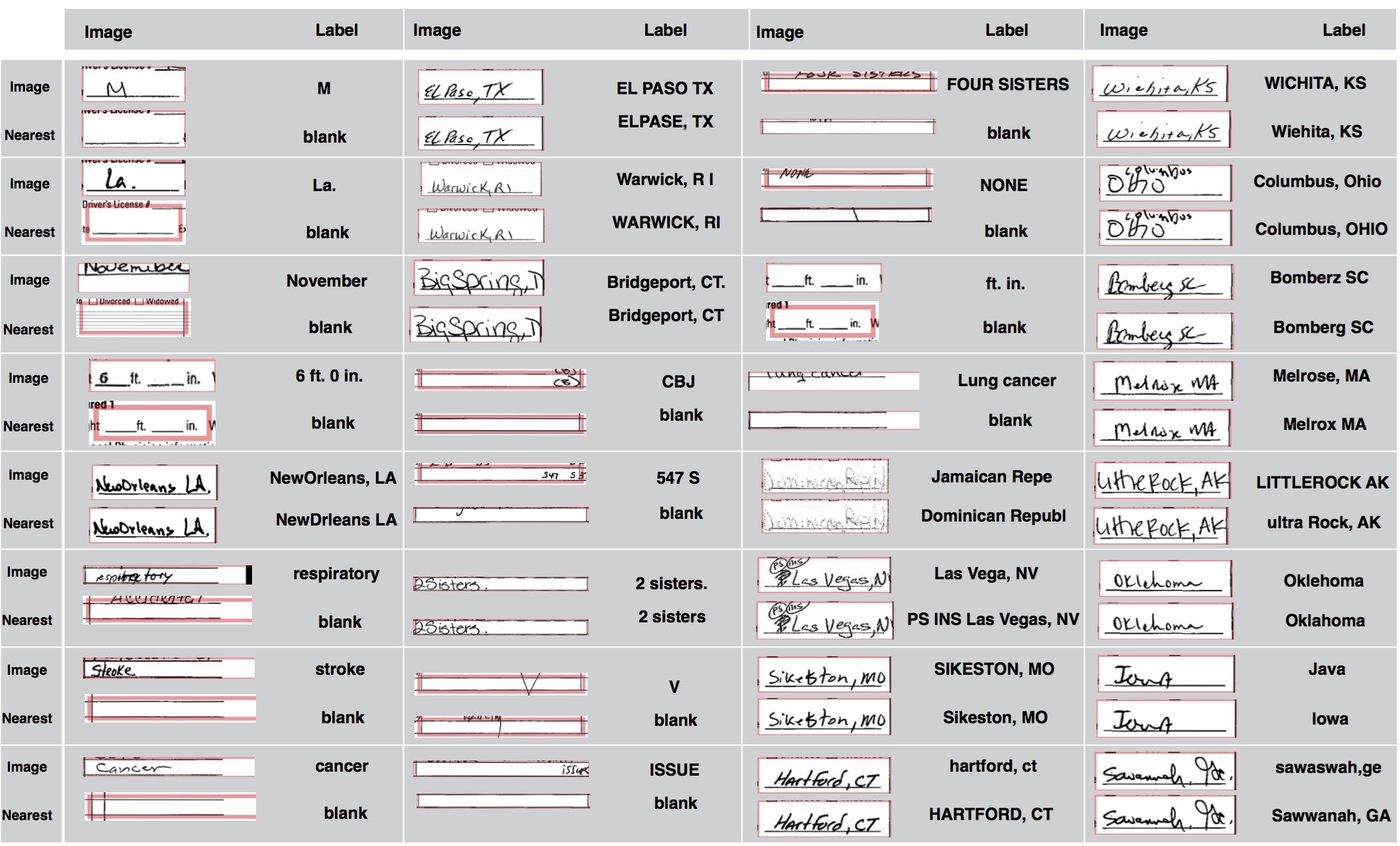}
		\caption{Texts with HCFN error (where DSSN-KNN produce high confidence wrong prediction). The nearest neighbor text in Similarity Manifold chosen by KNN is shown.}
		\label{fig:exp-err}
	\end{figure}

	 Some of the text images where DSSN-KNN produces high confidence errors are shown in Fig.~\ref{fig:exp-err}.  It is evident that most of the example pairs are, in fact, mutually visually similar, and the "errors" can be attributed to human errors in ground truth.  Interestingly, DSSN-KNN sometimes predicts better-than-human labels, for example, spelling corrections.   
	
	\vspace{-3mm}	
	\section{Conclusion}
	\vspace{-3mm}
	In this paper, we proposed a new text recognition model based on visual similarity of text images. A Deeply Supervised Siamese Network is trained along with a K-nearest neighbor classifier, to predict labels of text images. The performance of the proposed model is evaluated for accuracy and reduction of human cost of labeling. The results show that the average value of human-less efficiency on successful field is: $~25-45\%$ in ASSISTIVE mode with NO error, and $~50-85\%$ in ROBOTIC mode with $<0.5\%$ error.  Observed errors are explainable.  Predicted labels are sometimes better than human labels e.g. spell corrections.  Some of the false negative errors we count are in whitespace and irrelevant punctuation (the "real" error is lower than reported here).
	
	\bibliography{references}
	\bibliographystyle{iclr2016_conference}
	
\end{document}